\documentclass[10pt,twocolumn,letterpaper]{article}

\usepackage{amsmath,amsfonts,bm,amssymb,xspace}

\def\Figref#1{Figure~\ref{#1}}

\def\eqref#1{equation~\ref{#1}}
\def\1{\bm{1}}

\def\vmu{{\bm{\mu}}}
\def\vtheta{{\bm{\theta}}}

\def\vp{{\bm{p}}}

\def\vw{{\bm{w}}}

\def\vz{{\bm{z}}}

\def\vtheta{{\bm \theta}}

\def\vmu{{\bm \mu}}

\def\mM{{\bm{M}}}

\def\mP{{\bm{P}}}

\def\mW{{\bm{W}}}
\def\mX{{\bm{X}}}

\DeclareMathAlphabet{\mathsfit}{\encodingdefault}{\sfdefault}{m}{sl}
\SetMathAlphabet{\mathsfit}{bold}{\encodingdefault}{\sfdefault}{bx}{n}

\newcommand{\E}{\mathbb{E}}

\newcommand{\Tabref}[1]{Table \ref{#1}}

\newif\ifdraft
\ifdraft
    \usepackage[colorinlistoftodos]{todonotes}
    \newcommand{\todoline}[1]{\todo[inline]{#1}}
\else
    \newcommand{\todoline}[1]{}
\fi

\usepackage{cvpr}
\usepackage{times}
\usepackage{epsfig}
\usepackage{graphicx}
\usepackage{xspace,multirow}

\usepackage{algorithm}
\usepackage[noend]{algpseudocode}
\algrenewcommand\algorithmicdo{:}

\newcommand{\autoaugment}{AutoAugment\xspace}
\newcommand{\faster}{Faster \autoaugment}
\newcommand{\fast}{Fast \autoaugment}

\usepackage[breaklinks=true,bookmarks=false]{hyperref}

\cvprfinalcopy

\def\cvprPaperID{} % *** Enter the CVPR Paper ID here

\ifcvprfinal\fi
\begin{document}

%%%%%%%%% TITLE
\title{\faster: Learning Augmentation Strategies using Backpropagation}

\author{Ryuichiro Hataya${}^{1,2}$ \and Jan Zdenek${}^{1}$ \and Kazuki Yoshizoe${}^{2}$ \and Hideki Nakayama${}^{1}$ \\
${}^{1}$ Graduate School of Information Science and Technology, The University of Tokyo, Tokyo, Japan\\
${}^{2}$ RIKEN Center for Advanced Intelligence Project, Tokyo, Japan\\
}

\maketitle

\begin{abstract}

Data augmentation methods are indispensable heuristics to boost the performance of deep neural networks, especially in image recognition tasks. Recently, several studies have shown that augmentation strategies found by search algorithms outperform hand-made strategies. Such methods employ black-box search algorithms over image transformations with continuous or discrete parameters and require a long time to obtain better strategies. In this paper,  we propose a differentiable policy search pipeline for data augmentation, which is much faster than previous methods. We introduce approximate gradients for several transformation operations with discrete parameters as well as the differentiable mechanism for selecting operations. As the objective of training, we minimize the distance between the distributions of augmented data and the original data, which can be differentiated. We show that our method, \faster, achieves significantly faster searching than prior work without a performance drop.

\end{abstract}

\section{Introduction}\label{sec:intorduction}

Data augmentation is a powerful technique for machine learning to virtually increase the amount and diversity of data, which improves the performance especially in image recognition tasks. Conventional data augmentation methods include geometric transformations such as rotation and color enhancing such as auto-contrast. 
Similarly to other hyper-parameters, the designers of data augmentation strategies usually select transformation operations based on their prior knowledge (e.g., required invariance). For example, horizontal flipping is expected to be effective for general object recognition but probably not for digit recognition. In addition to the selection, the designers need to combine several operations and set their magnitudes (e.g., degree of rotation). Therefore, designing of data augmentation strategies is a complex combinatorial problem.

\begin{figure}[tb]
    \centering
    \includegraphics[width=0.7\linewidth]{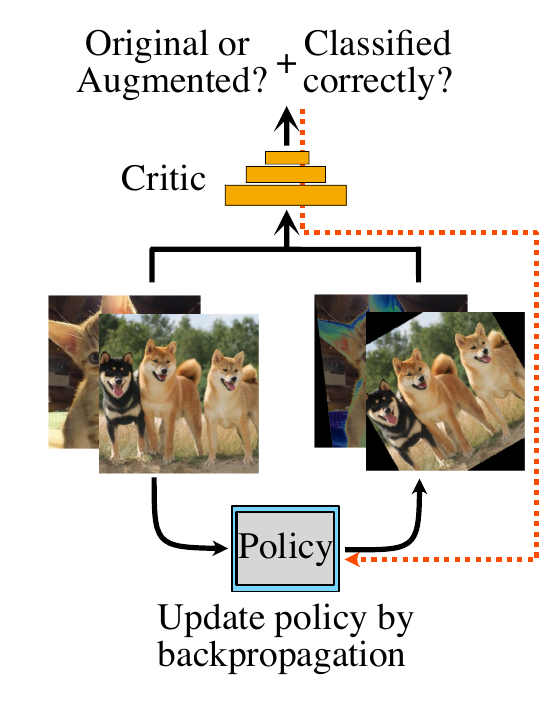}
    \caption{Overview of our proposed model. We propose to use a \textbf{differentiable data augmentation pipeline} to achieve faster policy search by using adversarial learning.}
    \label{fig:catch}
\end{figure}

\begin{table}[tb]
    \centering
    \begin{tabular}{c|c|c|c|c}
        Dataset & AA  & PBA  & Fast AA  & Faster AA (ours) \\ \hline
        CIFAR-10 & 5,000 & 5.0 & 3.5 & \textbf{0.23} \\ \hline
        SVHN & 1,000 & 1.0 & 1.5 & \textbf{0.061} \\ \hline
        ImageNet & 15,000 & - & 450 & \textbf{2.3}
    \end{tabular}
    \vspace{5pt}
    \caption{\textbf{\faster is faster than others, without a significant performance drop}  (see section \ref{sec:experiments}). GPU hours comparison of \faster (Faster AA), \autoaugment (AA) \cite{Cubuk2018}, PBA \cite{Ho2019} and \fast (Fast AA) \cite{Lim2019}.}
    \label{tab:gpuhours}
    \vspace{-5pt}
\end{table}

When designing data augmentation strategies in a data-driven manner, one can regard the problem as searching for optimal hyper-parameters in a search space, which becomes prohibitively large as the combinations get complex. Therefore, efficient methods are required to find optimal strategies. If gradient information of these hyper-parameters is available, they can be efficiently optimized by gradient descent \cite{pmlr-v37-maclaurin15}. However, the gradient information is usually difficult to obtain because some magnitude parameters are discrete, and the selection process of operations is non-differentiable. Therefore, previous research to automatically design data augmentation policies has used black-box optimization methods that require no gradient information. For example, \autoaugment \cite{Cubuk2018} used reinforcement learning.

\begin{figure}[tb]
    \centering
    \includegraphics[width=0.95\linewidth]{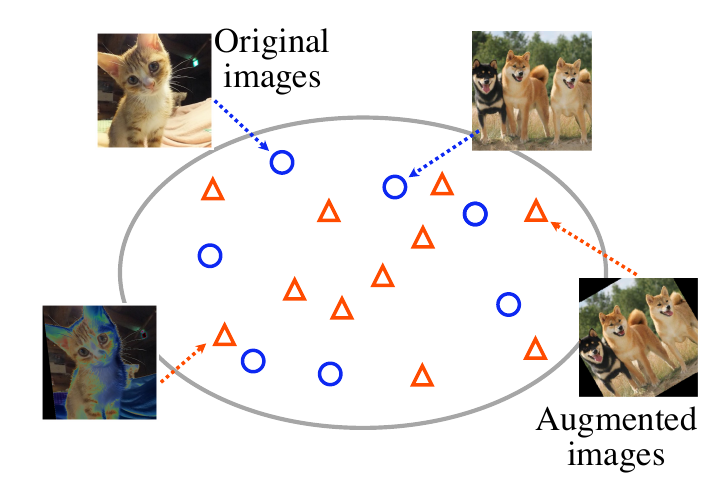}
    \caption{We regard data augmentation as a process that \textbf{fills missing data points of the original training data}; therefore, our objective is to minimize the distance between the distributions of augmented data and the original data using adversarial learning.}
    \label{fig:filling}
\end{figure}

In this paper, we propose to solve the problem by approximating gradient information and thus enabling gradient-based optimization for data augmentation policies. To this end, we approximate the gradients of discrete image operations using straight-through estimator \cite{Bengio2013} and make the selection process of operations differentiable by incorporating a recent differentiable neural architecture search method \cite{Liu2018c}. As the objective, we minimize the distance between the distributions of the original images and augmented images, because we want the data augmentation pipeline to transform images so that it fills missing points in the training data \cite{Lim2019} (see \Figref{fig:filling}). To make the transformed images match the distribution of original images, we use adversarial learning (see \Figref{fig:catch}). As a result, the searching process becomes end-to-end differentiable and significantly \textbf{faster} than prior work such as \autoaugment, PBA and \fast ~(see \Tabref{tab:gpuhours} \footnote{Note that \cite{Lim2019} and we estimate the GPU hours with an NVIDIA V100 GPU while \cite{Cubuk2018} did with an NVIDIA P100 GPU.}).

We empirically show that our method, which we call \faster, enables much faster policy search while achieving comparable performance with that of prior work on standard benchmarks: CIFAR-10, CIFAR-100 \cite{Krizhevsky2009}, SVHN \cite{Netzer2011} and ImageNet \cite{Russakovsky2015}.

In summary, our contributions are following three points:

\begin{enumerate}
    \item We introduce gradient approximations for several non-differentiable data augmentation operations.
    \item We make the searching of data augmentation policies end-to-end differentiable by gradient approximations, differentiable selection of operations and a differentiable objective that measures the distance between the original and augmented image distributions.
    \item We show that our proposed method, \faster, significantly reduces the searching time compared to prior methods without a performance drop.
\end{enumerate}

\section{Related Work}
\subsection*{Neural Architecture Search}

Neural Architecture Search (NAS) aims to automatically design architectures of neural networks to achieve higher performance than manually designed ones. To this end, NAS algorithms are required to select better combinations of components (e.g., convolution with a 3x3 kernel) from discrete search spaces using searching algorithms such as reinforcement learning  \cite{Zoph2016b} and evolution strategy \cite{Real2019}. Recently, DARTS \cite{Liu2018c} achieved faster search by relaxing the discrete search space to a continuous one which allowed them to use gradient-based optimization. While \autoaugment \cite{Cubuk2018} was inspired by \cite{Zoph2016b}, our method is influenced by DARTS \cite{Liu2018c}.

\subsection*{Data Augmentation}

Data augmentation methods improve the performance of learnable models by increasing the virtual size and diversity of training data without collecting additional data samples. Traditionally, geometric transformations and color enhancing transformations have been used in image recognition tasks. For example, \cite{Krizhevsky2012,He2016b}  randomly apply horizontal flipping and cropping as well as alternation of image hues. In recent years, other image manipulation methods have been shown to be effective. \cite{zhong2017random, DeVries2017} cut out a random patch from the image and replace it with random noise or a constant value. Another strategy is to mix multiple images of different classes either by convex combinations \cite{Zhang2017c,Tokozume2017b} or by creating a patchwork from them \cite{yun2019cutmix}. In these studies, the selection of operations, their magnitudes and the probabilities to be applied are carefully hand-designed.

\subsection*{Automating Data Augmentation}

Similar to NAS, it is a natural direction to aim to automate data augmentation. One direction is to search for better combinations of symbolic operations using black-box optimization techniques: reinforcement learning \cite{Cubuk2018,Ratner2017a}, evolution strategy \cite{Volpi2018}, Bayesian optimization \cite{Lim2019} and Population Based Training \cite{Ho2019}. As the objective, \cite{Cubuk2018,Volpi2018,Ho2019} directly aim to minimize error rate, or equivalently to maximize accuracy, while \cite{Ratner2017a,Lim2019} try to match the densities of augmented and original images.

Another direction is to use generative adversarial networks (GANs) \cite{Goodfellow2014b}. \cite{Tran2017a,Antoniou2019a} use conditional GANs to generate images that promote the performance of image classifiers. \cite{Shrivastava2017a,Sixt2018} use GANs to modify the outputs of simulators to look like real objects. 

Automating data augmentation can also be applied to representation learning such as semi-supervised learning \cite{Berthelot2019,Xie2019} and domain generalization \cite{Volpi2018}.

\section{Preliminaries}

\begin{figure}
    \centering
    \includegraphics[width=0.9\linewidth]{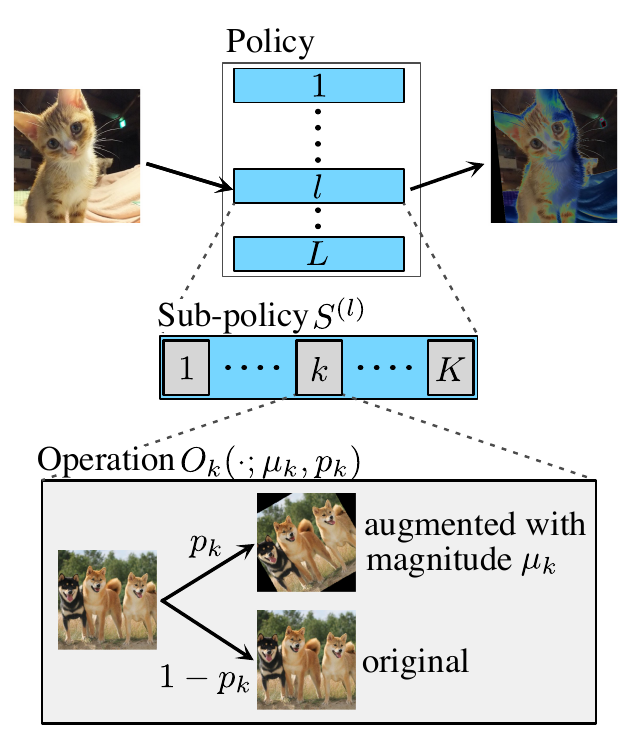}
    \caption{Schematic view of the problem setting. Each image is augmented by a \textbf{sub-policy} randomly selected from the \textbf{policy}. A single sub-policy is composed of $K$ consecutive \textbf{operations} $(O_1, \dots, O_K)$, such as \texttt{shear\_x} and \texttt{solarize}. An operation $O_k$ operates a given image with probability $p_k$ and magnitude $\mu_k$.}
    \label{fig:overview}
\end{figure}

In this section, we describe the common basis of \autoaugment \cite{Cubuk2018}, PBA \cite{Ho2019} and \fast \cite{Lim2019} (see also \Figref{fig:overview}). \faster also follows this problem setting.

In these works, input images are augmented by a policy which consists of $L$ different sub-policies $S^{(l)} ~(l=1,2,\dots,L)$. A randomly selected sub-policy transforms each image $\mX$. A single sub-policy consists of $K$ consecutive image processing operations $O_1^{(l)},\dots,O_K^{(l)}$ which are applied to the image one by one. We refer to the number of consecutive operations $K$ as operation count. In the rest of this paper, we focus on sub-policies; therefore, we omit the superscripts $l$.

Each method first searches for better policies. After the searching phase, the obtained policy is used as a data augmentation pipeline to train neural networks.

\subsection{Operations}\label{sub:operations}

\begin{table}[tb]
    \centering
    \begin{tabular}{l|l|l}
           & Operation & Magnitude $\mu$ \\ \hline
       \multirow{5}{*}{\shortstack[l]{Affine\\transformation}}
             & \texttt{shear\_x} & continuous \\
             & \texttt{shear\_y} & continuous \\
             & \texttt{translate\_x} & continuous \\
             & \texttt{translate\_y} & continuous \\
             & \texttt{rotate} & continuous \\
             
             & \texttt{flip}   & none \\ \hline
       \multirow{9}{*}{\shortstack[l]{Color\\enhancing\\operations}}
             & \texttt{solarize} & discrete \\
             & \texttt{posterize} & discrete \\
             & \texttt{invert} & none \\
             & \texttt{contrast} & continuous \\
             & \texttt{color} & continuous \\
             & \texttt{brightness} & continuous \\
             & \texttt{sharpness} & none \\
             & \texttt{auto\_contrast} & none \\
             & \texttt{equalize}   & none \\ \hline
       \multirow{2}{*}{Other operations}
             & \texttt{cutout} & discrete \\
             & \texttt{sample\_pairing} & continuous
    \end{tabular}
    \vspace{5pt}
    \caption{Operations used in \autoaugment, PBA, \fast and \faster. Some operations have discrete magnitude parameters $\mu$, while others have no or continuous magnitude parameters. Different from previous works, we approximate gradients of operations w.r.t. discrete magnitude $\mu$, which we describe in section \ref{subsub:magnitude}.}
    \label{tab:operations}
\end{table}

Operations used in each sub-policy include affine transformations such as \texttt{shear\_x} and color enhancing operations such as \texttt{solarize}. In addition, we use \texttt{cutout} \cite{DeVries2017} and \texttt{sample\_pairing} \cite{Inoue} following \cite{Cubuk2018,Ho2019,Lim2019}. We show all 16 operations used in these works in \Tabref{tab:operations}. We denote the set of operations as $\mathcal{O}=\{\mathtt{shear\_x},\mathtt{solarize},\ldots\}$.

Some operations have magnitude parameters that are free variables, e.g., the angle in \texttt{rotate}. On the other hand, some operations, such as \texttt{invert}, have no magnitude parameter. For simplicity, we use the following expressions as if every operation had its magnitude parameter $\mu_O  (\in[0, 1])$. Each operation is applied with probability of $p_O (\in[0, 1])$. Therefore, each image $\mX$ is augmented as

\begin{equation}\label{eq:operation}
    \mX \to \begin{cases}
         O(\mX; \mu_O) & ~~(\text{with probability of } p_O) \\
         \mX & ~~(\text{with probability of } 1-p_O).
    \end{cases}
\end{equation}

Rewriting this mapping as $O(\cdot; \mu_O, p_O)$, each sub-policy $S$ consisting of operations $O_1, O_2, \dots, O_K$ can be written as

\begin{equation}\label{eq:subpolicy}
    S(\mX; \vmu_S, \vp_S)=(O_K\circ \dots \circ O_1)(\mX; \vmu_S, \vp_S),
\end{equation}

\noindent where $\vmu_S=(\mu_{O_1}, \dots, \mu_{O_K})$ and $\vp_S=(p_{O_1}, \dots, p_{O_K})$. In the rest of this paper, we represent an image operation $O$, $O(\cdot; \mu)$ and $O(\cdot; \mu, p)$ interchangeably according to the context.

\subsection{Search Space}\label{sub:search_space}

The goal of searching is to find the best operation combination $O_1, \dots, O_K$ and parameter sets $(\vmu_S, \vp_S)$ for $L$ sub-policies. Therefore, the size of the total search space is roughly $(\#\mathcal{O}\times[0, 1]\times[0, 1])^{KL}$. Using multiple sub-policies results in a prohibitively large search space for brute-force searching. \cite{Lim2019} uses Bayesian optimization in this search space. \cite{Cubuk2018,Ho2019} discretize the continuous part $[0, 1]$ into $10$ or $11$ values and search the space using reinforcement learning and population based training. Nevertheless, the problem is still difficult to solve naively even after discretizing the search space. For instance, if the number of sub-policies $L$ is $10$ with $K=2$ consecutive operations, the discretized space size becomes $(16\times 10\times 11)^{2\times 10}\approx 8.1\times 10^{64}$. 

Previous methods \cite{Cubuk2018,Ho2019,Lim2019} use black-box optimization. Therefore, they need to train CNNs with candidate policies and obtain their validation accuracy. The repetition of this process requires a lot of time. In contrast, \faster achieves faster searching with gradient-based optimization to avoid repetitive evaluations, even though the search space is the same as in \fast. We describe the details in the next section.

\section{\faster}\label{sec:faster_auto_augment}

\faster explores the search space to find better policies in a gradient-based manner, which distinguishes our method. In section \ref{sub:differentiable_da}, we describe the details of gradient approximation for policy searching. To accomplish gradient-based training, we adopt distance minimization between the distributions of the augmented and the original images as the learning objective, which we present in section \ref{sub:density_matching}.

\subsection{Differentiable Data Augmentation Pipeline}\label{sub:differentiable_da}

Previous searching methods \cite{Cubuk2018,Ho2019,Lim2019} have used image processing libraries (e.g., Pillow) which do not support backpropagation through the operations in \Tabref{tab:operations}. Contrary to previous work, we modify these operations to be differentiable --- each of which can be differentiated with respect to the probability $p$ and the magnitude $\mu$. Thanks to this modification, the searching problem becomes an optimization problem. The sequence of operations in each sub-policy also needs to be optimized in the same fashion.

\subsubsection*{On the probability parameter $p$}

First, we regard \eqref{eq:operation} as 

\begin{equation}
    bO(\mX; \mu)+(1-b)\mX,
\end{equation}

\noindent where $b \in\{0, 1\}$ is sampled from Bernoulli distribution $\mathrm{Bern}(b; p)$, i.e. $b=1$ with probability of $p$. Since this distribution is non-differentiable, we instead use Relaxed Bernoulli distribution \cite{Jang2016}

\begin{equation}
    \mathrm{ReBern}(b; p, \lambda)=\varsigma(\frac{1}{\lambda}\{\log\frac{p}{1-p}+\log\frac{u}{1-u}\}).
\end{equation}

Here, $\displaystyle \varsigma(x)=\frac{1}{1+\exp(-x)}$ is a sigmoid function that keeps the range of function in $(0, 1)$ and $u$ is a value sampled from a uniform distribution on $[0, 1]$. With low temperature of $\lambda$, this relaxed distribution behaves like Bernoulli distribution. Using this reparameterization, each operation $O(\cdot; \mu_O, p_O)$ can be differentiable w.r.t. its probability parameter $p$.

\subsubsection*{On the magnitude parameter $\mu$}\label{subsub:magnitude}

For some operations, such as \texttt{rotate} or \texttt{translate\_x}, their gradients w.r.t. their magnitude parameters $\mu$ can be obtained easily. However, some operations such as \texttt{posterize} and \texttt{solarize} discretize magnitude values. In such cases, gradients w.r.t. $\mu$ cannot backpropagate through these operations. Thus, we approximate their gradient in a similar manner to the straight-through estimator \cite{Bengio2013,Oord2017}. More precisely, we approximate the $(i, j)$th element of an augmented image by an operator $O$ as

\begin{equation}
    \tilde{O}(\mX; \mu)_{i,j} = \mathrm{StopGrad}(O(\mX; \mu)_{i,j}-\mu) + \mu, \label{eq:straight_through}
\end{equation}

\noindent where $\mathrm{StopGrad}$ is a stop gradient operation which treats its operand as a constant. During the forward computation, the augmentation is exactly operated: $\tilde{O}(\mX; \mu)_{i,j}=O(\mX; \mu)_{i,j}$. However, during the backward computation, the first term of the right-hand side of equation \ref{eq:straight_through} is ignored because it is constant, and then we obtain an approximated gradient:

\begin{equation}
 \frac{\partial O(\mX)_{i,j}}{\partial \mu}\approx\frac{\partial \tilde{O}(\mX)_{i,j}}{\partial \mu}= 1.
\end{equation}

Despite its simplicity, we find that this method works well in our experiments. Using this approximation, each operation $O(\cdot; \mu_O, p_O)$ can be differentiable w.r.t. its magnitude parameter $\mu$.

\subsubsection*{Searching for operations in sub-policies}

\begin{figure}[tb]
    \centering
    \includegraphics[width=0.9\linewidth]{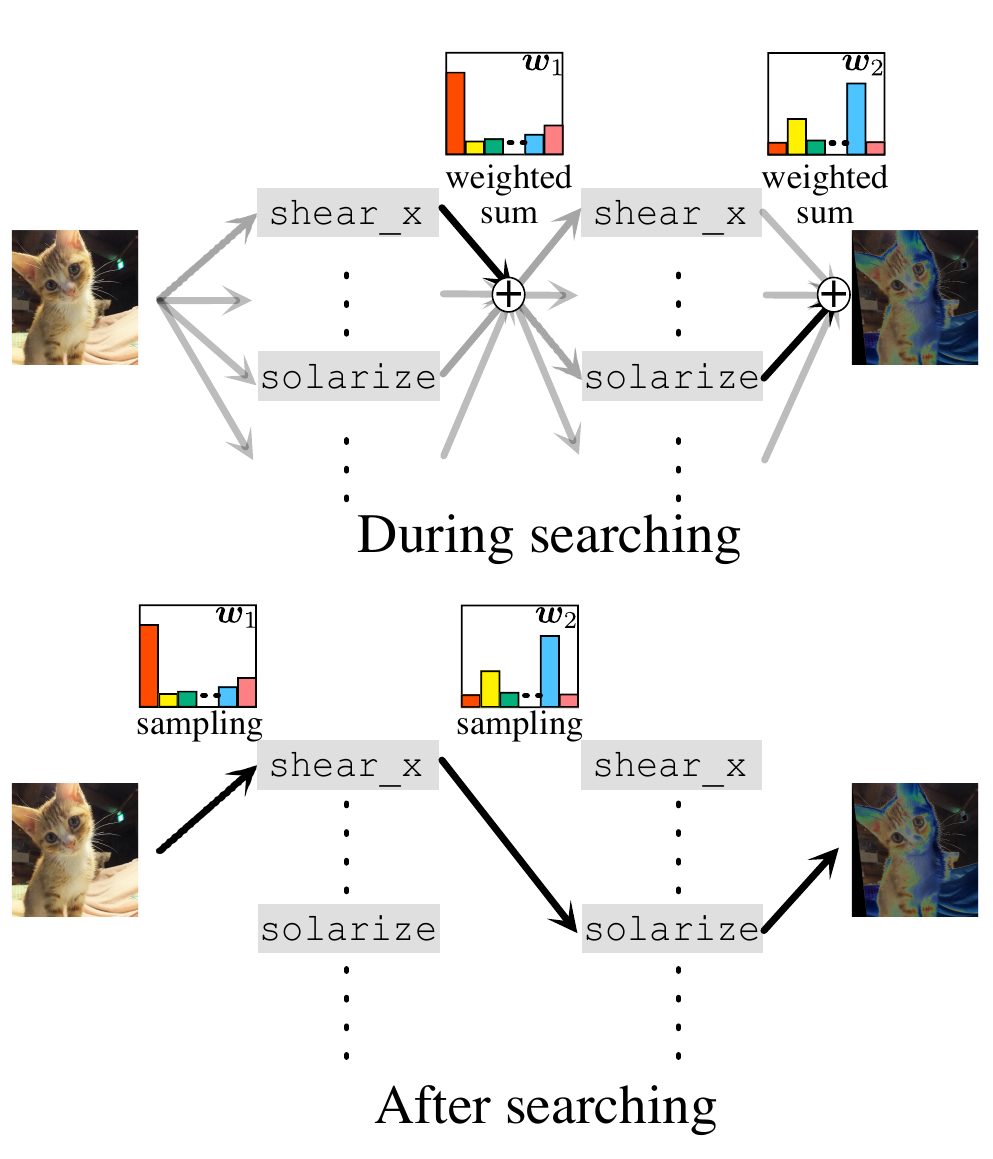}
    \caption{Schematic view of the selection of operations in a single sub-policy when $K=2$. During searching, we apply all operations to an image and take weighted sum of the results as an augmented image. The weights, $\vw_1$ and $\vw_2$, are also updated as other parameters. After searching, we sample operations according to the trained weights.}
    \label{fig:soft_path}
\end{figure}
                        
Each sub-policy $S$ consists of $K$ operations. To select the appropriate operation $O_k$ where $k \in \{1,2,\dots,K\}$, we use a strategy similar to the one used in neural architecture search \cite{Liu2018c} (see also Algorithm \ref{alg:combination} and \Figref{fig:soft_path} for details). To be specific, we approximate the output of a single selected $k$th operation $O_k(\mX)$ by weighted sum of the outputs of all operations as 

\begin{equation}
  \sum_n^{\#\mathcal{O}} [\sigma_\eta(\vw_k)]_n O_k^{(n)}(\mX ;\mu_k^{(n)}, p_k^{(n)}),
\end{equation}

\noindent where, $O_k^{(n)}$ is an operation in $\mathcal{O}$, and $O_k^{(n)}$ and $O_k^{(m)}$ are different operations if $n\neq m$. $\vw_k$ is a learnable parameter and $\sigma_\eta$ is a softmax function $\displaystyle \sigma_\eta(\vz)=\frac{\exp(\vz/\eta)}{\sum_j \exp(z_i/\eta)}$ with a temperature parameter $\eta>0$. With a low temperature $\eta$, $\sigma_\eta(\vw_k)$ becomes a onehot-like vector. During inference, we sample the $k$th operation according to categorical distribution $\mathrm{Cat}(\sigma_k(\vw_k))$.

\begin{algorithm}
\caption{Selection of operations in a single sub-policy during searching. Refer to \Figref{fig:soft_path} for the  $K=2$ case.}
\label{alg:combination}

\begin{algorithmic}
    \Statex $\mX$: input image, 
    \Statex $\{\vw_1,\dots, \vw_K\}$: learnable weights, 
    \Statex $\sigma_\eta$: softmax function with temperature $\eta$
    \For{$k$ in $\{1, 2, \dots, K\}$}
        \State Augment $\mX$ by the $k$th stage operations: 
        
        $\displaystyle \mX\leftarrow\sum_{n=1}^{\#\mathcal{O}} [\sigma_\eta(\vw_k)]_nO_k^{(n)}(\mX; \mu_k^{(n)}, p_k^{(n)})$
    \EndFor
    \Return $\mX$
\end{algorithmic}
\end{algorithm}

\subsection{Data Augmentation as Density Matching}\label{sub:density_matching}

Using the techniques described above, we can back-propagate through the data augmentation process. In this section, we describe the objective of policy learning.

One possible candidate for the objective is the minimization of the validation loss as in DARTS \cite{Liu2018c}. However, this bi-level formulation takes a lot of time and costs a large memory footprint \cite{Finn2017b}. To avoid this problem, we adopt a different approach.

Data augmentation can be seen as a process that fills missing data points in training data \cite{Lim2019,Ratner2017a,Tran2017a}. Therefore, we minimize the distance between distributions of the original images and the augmented images. This goal can be achieved by minimizing the Wasserstein distance between these distributions $d_\vtheta$ using Wasserstein GAN \cite{Arjovsky2017b} with gradient penalty \cite{Gulrajani2017}. Here, $\vtheta$ is the parameters of its critic, or almost equivalently, discriminator. 
Unlike usual GANs for image modification, our model does not have a typical generator that learns to transform images using conventional neural network layers. Instead, a policy --- explained in previous sections --- is trained, and it transforms images using predefined operations.
Following prior work \cite{Cubuk2018,Ho2019,Lim2019}, we use WideResNet-40-2 \cite{Zagoruyko} (for CIFAR-10, CIFAR-100 and SVHN) or ResNet-50 \cite{He2016b} (for ImageNet) and replace their classifier heads with a two-layer perceptron  that serves as a critic. Besides, we add a classification loss to prevent images of a certain class to be transformed into images of another class (see Algorithm \ref{alg:faster_autoaugment}).

\begin{algorithm}
\caption{Training of \faster}
\label{alg:faster_autoaugment}

\begin{algorithmic}
    \Statex $\mM, \mP, \mW$: learnable parameters of a sub-policy
    \Statex $d_\vtheta(\cdot, \cdot)$: distance between two densities with learnable parameters $\vtheta$ 
    \Statex $f$: image classifier
    \Statex $\mathcal{L}$: cross entropy loss
    \Statex $\epsilon$: coefficient of classification loss
    \Statex $\mathcal{D}$: training set
    \While{not converge}
        \State Sample a pair of batches $\mathcal{B}, \mathcal{B}'$ from $\mathcal{D}$
        \State Augment data $\mathcal{A}=\{S(\mX;\mM, \mP, \mW); (\mX, \cdot)\in\mathcal{B}\}$
        \State Measure distance $d=d_\vtheta(\mathcal{A}, \mathcal{B}')$
        \State Classification loss \\ $l=\E_{(\mX, y)\sim\mathcal{A}}\mathcal{L}(f(\mX), y)+\E_{(\mX', y')\sim\mathcal{B}'}\mathcal{L}(f(\mX'), y')$
        \State Update parameters $\mM, \mP, \mW, \vtheta$ to minimize $d+\epsilon l$ using stochastic gradient descent (e.g., Adam)
    \EndWhile
\end{algorithmic}
\end{algorithm}

\section{Experiments and Results}\label{sec:experiments}

In this section, we show the empirical results of our approach on CIFAR-10, CIFAR-100 \cite{Krizhevsky2009}, SVHN \cite{Netzer2011} and ImageNet \cite{Russakovsky2015} datasets and compare the results with \autoaugment \cite{Cubuk2018}, PBA \cite{Ho2019} and \fast \cite{Lim2019}. Except for ImageNet, we run all experiments three times and report the average results. The details of datasets are presented in \Tabref{tab:dataset_summary}.

\subsection{Implementation Details}

Prior methods \cite{Cubuk2018,Ho2019,Lim2019} employed Python's Pillow \footnote{\url{https://python-pillow.org/}} as the image processing library. We transplanted the operations described in section \ref{sub:operations} to PyTorch \cite{Paszke2017}, a tensor computation library with automatic differentiation. For geometric operations, we extend functions in kornia \cite{eriba2019kornia}. For color-enhancing operations, sample pairing \cite{Inoue} and cutout \cite{DeVries2017}, we implement them using PyTorch. Operations with discrete magnitude parameters are implemented as described in section \ref{sub:differentiable_da} with additional CUDA kernels.

We use CNN models and baseline preprocessing procedures available from the \fast's repository \footnote{\url{https://github.com/kakaobrain/fast-autoaugment/tree/master/FastAutoAugment/networks}} and follow their settings and hyper-parameters for CNN training such as the initial learning rate and learning rate scheduling.

\subsection{Experimental Settings}\label{sub:experimental_settings}

To compare our results with previous studies \cite{Cubuk2018,Lim2019,Ho2019}, we follow their experimental settings on each dataset. We train the policy on randomly selected subsets of each dataset presented in \Tabref{tab:dataset_summary}. In the evaluation phase, we train CNN models from scratch on each dataset with learned \faster policies. For SVHN, we use both training and additional datasets.

Similar to \fast \cite{Lim2019}, our policies are composed of 10 sub-policies each of which has operation count $K=2$ as described in section \ref{sub:search_space}. We train the policies for 20 epochs using ResNet-50 for ImageNet and WideResNet-40-2 for other datasets. In all experiments, we set temperature parameters $\lambda$ and $\eta$ to $0.05$. We use Adam optimizer \cite{Kingma2015} with a learning rate of $1.0^{-3}$, coefficients for running averages (betas) of $(0, 0.999)$, the coefficient for the classification loss $\epsilon$ of $0.1$, and the coefficient for gradient penalty of $10$. Because GPUs are optimized for batched tensor computation, we apply sub-policies to chunks of images. The number of chunks determines the balance between speed and diversity. We set the chunk size to $16$ for ImageNet and $8$ for other datasets during searching. For evaluation, we use the chunk size of $32$ for ImageNet and $16$ for other datasets.

\begin{table}[tb]
    \centering
    
    \resizebox{\columnwidth}{!}{%
    \begin{tabular}{l|r|r}
       Dataset  & Training set size & Subset size for policy training  \\ \hline
       CIFAR-10 \cite{Krizhevsky2009} & 50,000 & 4,000 \\
       CIFAR-100 \cite{Krizhevsky2009} & 50,000 & 4,000 \\
       SVHN \cite{Netzer2011} & 603,000 & 1,000 \\
       ImageNet \cite{Russakovsky2015} &  1,200,000 & 6,000
    \end{tabular}
    }
    \vspace{5pt}
    \caption{Summary of datasets used in the experiments. For the policy training on ImageNet, we use only 6000 images from the 120 selected classes following \cite{Cubuk2018,Lim2019}.}
    \label{tab:dataset_summary}
\end{table}

\subsection{Results}\label{sub:qualititative_results}

\subsubsection*{CIFAR-10 and CIFAR-100}

In \Tabref{tab:misc_results}, we show test error rates on CIFAR-10 and CIFAR-100 with various CNN models: WideResNet-40-2, WideResNet-28-10 \cite{Zagoruyko}, Shake-Shake ($26\ 2\times\{32, 96, 112\}$d) \cite{Gastaldi}. We train WideResNets for 200 epochs and Shake-Shakes for 1,800 epochs as \cite{Cubuk2018} and report averaged values over three runs for \faster. The results of baseline and Cutout are from \cite{Cubuk2018,Lim2019}.  \faster not only shows competitive results with prior work, but this method is significantly faster to train than others (See \Tabref{tab:gpuhours}). For CIFAR-100, we report results with policies trained on reduced CIFAR-10 following \cite{Cubuk2018} as well as policies trained on reduced CIFAR-100. The latter results are better than the former ones, which suggests the importance of training policy on the target dataset.

\begin{figure*}
    \centering
    \includegraphics[width=0.68\linewidth]{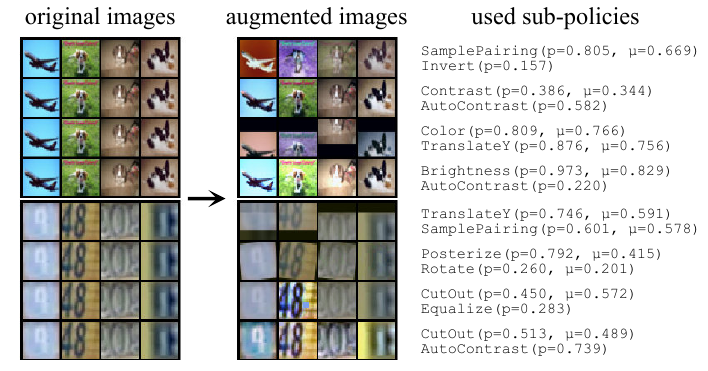}
    \caption{Original and augmented images of CIFAR-10 (upper) and SVHN (lower). As can been seen, \faster can transform original images into diverse augmented images with sub-policies at the right-hand side.}
    \label{fig:augmented_images}
\end{figure*}

We also show several examples of augmented images in \Figref{fig:augmented_images}. The policy seems to prefer color enhancing operations as reported in \autoaugment \cite{Cubuk2018}.

In \Tabref{tab:reduced_cifar10_results}\footnote{\cite{Cubuk2018} reports better baseline and Cutout performance than us  (18.8 \% and 16.5 \% respectively), but we could not reproduce the results.}, we report error rates on Reduced CIFAR-10 to show the effect of \faster in the low-resource scenario. In this experiment, we randomly sample 4,000 images from the training dataset. We train the policy using the subset and evaluate the policy with WideResNet-28-10 on the same subset for 200 epochs. As can be seen, \faster improves the performance 7.7 \% over Cutout and achieves a close error rate to \autoaugment. This result implies that data augmentation can moderately unburden the difficulty of learning from small data.

\begin{table*}[tb]
    \centering
    \resizebox{\linewidth}{!}{%
    \begin{tabular}{l|l|ccccc|c}
       Dataset & Model & Baseline & Cutout \cite{DeVries2017} & AA \cite{Cubuk2018} & PBA \cite{Ho2019} & Fast AA \cite{Lim2019} & Faster AA (ours) \\ \hline
       \multirow{5}{*}{CIFAR-10} &
        WideResNet-40-2 \cite{Zagoruyko}  &              5.3 & 4.1 & 3.7 & -   & 3.6 & 3.7 \\
       &WideResNet-28-10 \cite{Zagoruyko} &              3.9 & 3.1 & 2.6 & 2.6 & 2.7 & 2.6 \\
       &Shake-Shake (26 $2\times32$d) \cite{Gastaldi} & 3.6 & 3.0 & 2.5 & 2.5 & 2.7 & 2.7 \\
       &Shake-Shake (26 $2\times96$d) \cite{Gastaldi} & 2.9 & 2.6 & 1.9 & 2.0 & 2.0 & 2.0 \\
       &Shake-Shake (26 $2\times112$d) \cite{Gastaldi} & 2.8 & 2.6 & 1.9 & 2.0 & 2.0 & 2.0 \\ \hline
       
       \multirow{3}{*}{CIFAR-100} &
        WideResNet-40-2  \cite{Zagoruyko} &              26.0 & 25.2 & 20.7 & -    & 20.7 & 22.1 / 21.4 \\
       &WideResNet-28-10 \cite{Zagoruyko} &              18.8 & 18.4 & 17.1 & 16.7 & 17.3 & 17.8 / 17.3 \\
       &Shake-Shake (26 $2\times96$d) \cite{Gastaldi}  &17.1 & 16.0 & 14.3 & 15.3 & 14.9 & 15.6 / 15.0 \\ \hline
       
      SVHN & WideResNet-28-10 \cite{Zagoruyko} &   1.5 & 1.3 & 1.1 & 1.2 & 1.1 & 1.2 \\
    \end{tabular}
    }
    \vspace{5pt}
    \caption{\textbf{\faster yields comparable performance with prior work}. Test error rates on CIFAR-10, CIFAR-100 and SVHN. We report average rates over three runs. For CIFAR-100, we report results obtained with policies trained on CIFAR-10 / CIFAR-100.}
    \label{tab:misc_results}

\end{table*}

\begin{table}
    \centering

    \begin{tabular}{ccc|c}
       Baseline & Cutout \cite{DeVries2017} & AA \cite{Cubuk2018}    & Faster AA  (ours)  \\ \hline
       24.3 & 22.5 & 14.1 & 14.8
    \end{tabular}
    \vspace{5pt}
    \caption{Test error rates with models trained on Reduced CIFAR-10, which consists of 4,000 images randomly sampled from the training set. We show that the obtained policy by \faster is useful for the low-resource scenario.}
    \label{tab:reduced_cifar10_results}
    
\end{table}

\subsubsection*{SVHN}

In \Tabref{tab:misc_results}, we show test error rates on SVHN with WideResNet-28-10 trained for 200 epochs. For \faster, we report the average value of three runs. \faster achieves the error rate of $1.2\%$, which is $0.1\%$ improvement over Cutout and on par with PBA. The augmented images are seen in \Figref{fig:augmented_images}. Besides, we show the augmented images in \Figref{fig:augmented_images} with used sub-policies, which seem to select more geometric transformations than CIFAR-10's policy as reported in \autoaugment \cite{Cubuk2018}.

\subsubsection*{ImageNet}

In \Tabref{tab:imagenet_results}, we compare the top-1 and top-5 validation error rates on ImageNet with \cite{Cubuk2018,Lim2019}. To align our results with \cite{Cubuk2018}, we also train ResNet-50 for 200 epochs. \cite{Cubuk2018,Lim2019} report top-1 / top-5 error rates of $23.7$\% / $6.9$\%; however, despite efforts to reproduce the results, we could not reach the same baseline performance. \faster achieves a $1.0$\% improvement over the baseline on top-1 error rate. This gain is close to that of \autoaugment and \fast, which verifies that \faster has an effect comparable to prior work on a large and complex dataset.

\begin{table}
    \centering
    \begin{tabular}{lccc}
                           & Baseline & with Policy & Gain  \\ \hline
       AA \cite{Cubuk2018} &  23.7/6.9      &   22.4/6.2      &     1.3/0.7            \\
       Fast AA \cite{Lim2019} &       & 22.4/6.3      &       1.3/0.6          \\ \hline
       Faster AA (ours)  &   24.1/7.2        &  23.5/6.8      &     0.6/0.4             
    \end{tabular}
    \vspace{5pt}
    \caption{Top-1/Top-5 validation error rates on ImageNet \cite{Russakovsky2015} with ResNet-50 \cite{He2016b}. \faster achieves comparable performance gain to AA and Fast AA.}
    \label{tab:imagenet_results}
\end{table}

\section{Analysis}

\subsection{Changing the Number of Sub-policies}\label{sub:num_sub_policies}

\begin{figure*}[tb]
    \centering
    \begin{tabular}{c}
        \hspace{-0.02\linewidth}
        \begin{minipage}{0.48\linewidth}
            \includegraphics[width=\linewidth]{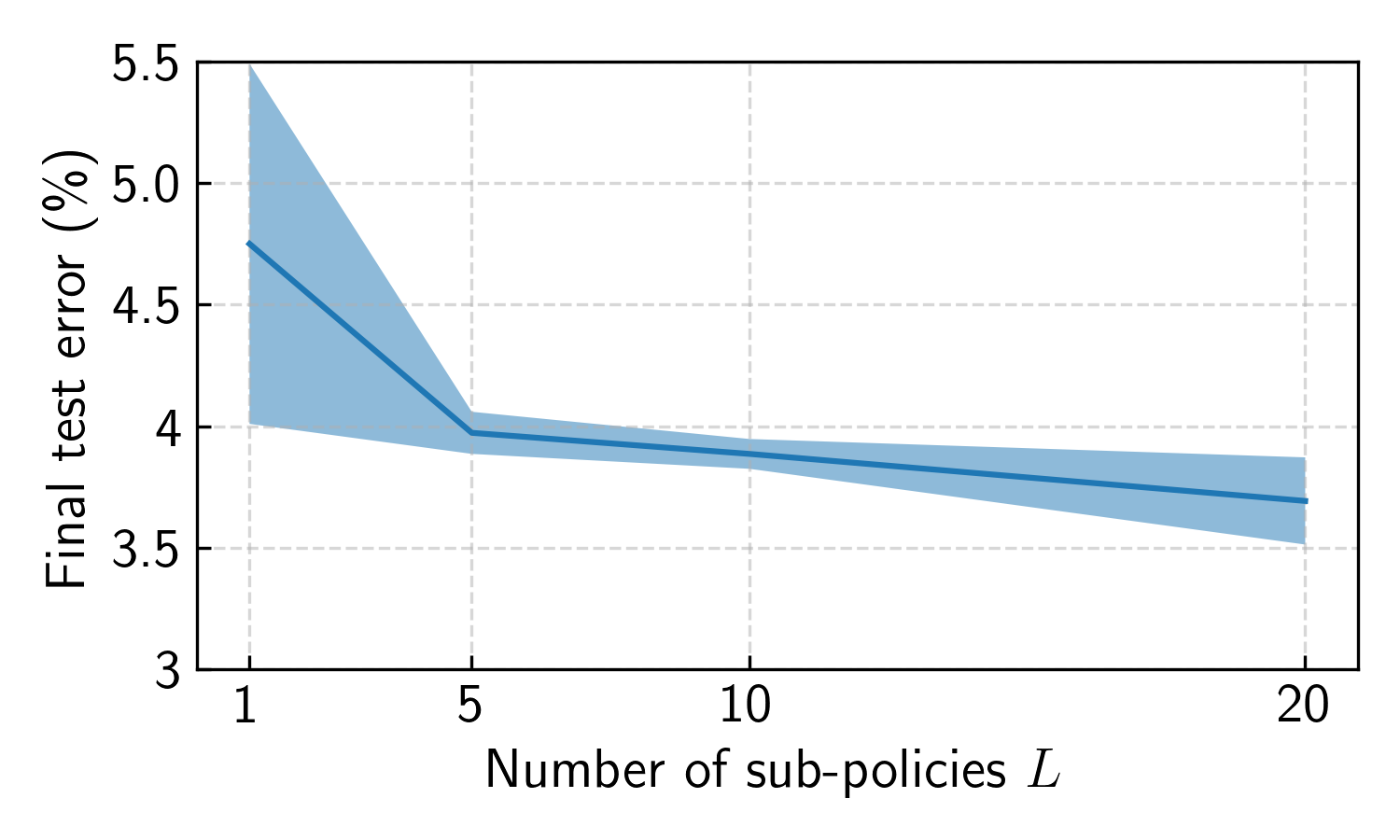}
            \caption{\textbf{As the number of sub-policies grows, performance increases.} The relationship between the number of sub-policies and the test error rate (CIFAR-10 with WideResNet-40-2). We plot test error rates and their standard deviations averaged over three runs.}
            \label{fig:num_sub_policies}
        \end{minipage}
        
        \hspace{0.02\linewidth}
        
        \begin{minipage}{0.48\linewidth}
            \vspace{-10pt}
            \includegraphics[width=\linewidth]{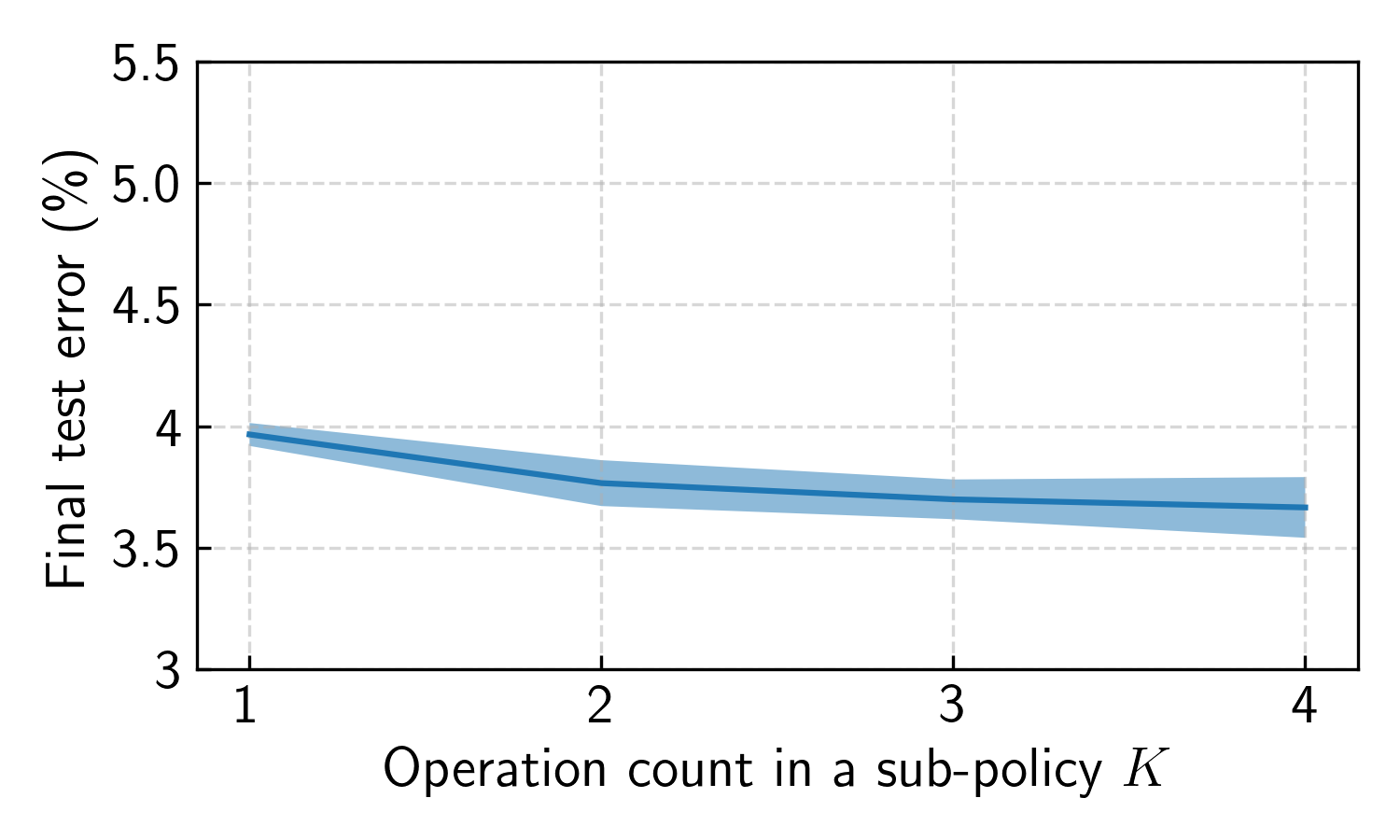}
            \caption{\textbf{As the operation count grows, performance increases.} The relationship between the operation count of each sub-policy and the average test error rate of three runs (CIFAR-10 with WideResNet-40-2).}
            \label{fig:num_operations}
        \end{minipage}
    \end{tabular}

\end{figure*}

The number of sub-policies $L$ is arbitrary. \Figref{fig:num_sub_policies} shows the relationship between the number of sub-policies and the final test error on CIFAR-10 dataset with WideResNet-40-2. As can be seen, the more sub-policies we have, the lower the error rate is. This phenomenon is straight-forward because the number of sub-policies determines the diversity of augmented images; however, an increase in the number of sub-policies results in exponential growth of the search space, which is prohibitive for standard searching methods.

\subsection{Changing Operation Count}\label{sub:operation_count}

The operation count $K$ of each sub-policy is also arbitrary. Like the number of sub-policies $L$, the operation count of a sub-policy $K$ also exponentially increases the search space. We change $K$ from $1$ to $4$ on CIFAR-10 dataset with WideResNet-40-2. We present the resulted error rates in \Figref{fig:num_operations}. As can be seen, as the operation count in each sub-policy grows, the performance increases, i.e., the error rates decrease. Results of section \ref{sub:num_sub_policies} and section \ref{sub:operation_count} show that \faster is scalable to a large search space.

\subsection{Changing Data Size}

In the main experiments in section \ref{sec:experiments}, we use a subset of CIFAR-10 of 4,000 images for policy training. To validate the effect of this sampling, we train a policy on the full CIFAR-10 of 50,000 images as \cite{Lim2019} and evaluate the obtained policy with WideResNet-40-2. We find that the increase of data size causes a significant performance drop (from $3.7$\% to $4.1$\%) with the number of sub-policies $L=10$. We hypothesize that this drop is because of lower capability of the policy when $L=10$. Therefore, we train a policy with $L=80$ sub-policies and randomly sample $10$ sub-policies to evaluate the policy, which results in comparable error rates ($3.8$\%). We present the results in \Tabref{tab:full_cifar10}, comparing with \fast \cite{Lim2019}, which shows the effectiveness of using subsets for \fast and \faster.

\begin{table}
    \centering
    \begin{tabular}{c|c|c}
    Data size & Fast AA \cite{Lim2019} & Faster AA (ours) \\ \hline
    4,000  & 3.6     & 3.7 \\
    50,000 & 3.7     &  3.8
    \end{tabular}
    \vspace{5pt}
    \caption{Test error rates on CIFAR-10 using policies trained on the reduced CIFAR-10 (4,000 images) and the full CIFAR-10 (50,000 images) with WideResNet-40-2.}
    \label{tab:full_cifar10}
\end{table}

\subsection{The Effect of Policy Training}

To confirm that trained policies are more effective than randomly initialized policies, we compare test error rates on CIFAR-10 with and without policy training, as performed in \autoaugment \cite{Cubuk2018}. Using WideResNet-28-10, trained policies achieve error rate of $2.6$\% while randomly initialized policies have a slightly worse error rate of $2.7$\% (both error rates are an average of three runs). These results imply that data augmentation policy searching is a meaningful research direction, but still has much room to improve.

\section{Conclusion}

In this paper, we have proposed \faster, which achieves faster policy searching for data augmentation than previous methods \cite{Cubuk2018,Ho2019,Lim2019}. To achieve this, we have introduced gradient approximation for several non-differentiable image operations and made the policy searching process end-to-end differentiable. We have verified our method on several standard benchmarks and showed that \faster could achieve competitive performance with other methods for automatic data augmentation. Besides, our additional experiments suggest that gradient-based policy optimization can scale to more complex scenarios.

We believe that faster policy searching will be beneficial for research on representation learning such as semi-supervised learning \cite{Berthelot2019,Xie2019} and domain generalization \cite{Volpi2018}. Additionally, learning from small data using learnable policies might be an interesting future direction.

\clearpage
{\small
\bibliographystyle{ieee}
\bibliography{ms}
}

\end{document}